\documentclass{article}

\usepackage{microtype}
\usepackage{graphicx}
\usepackage{subfigure}
\usepackage{booktabs} 
\usepackage{amsmath}
\usepackage{multirow}
\usepackage{enumitem}
\usepackage{booktabs}

\usepackage{hyperref}

\usepackage[accepted]{icml2021}

\icmltitlerunning{\TheModelsp: Deep Mixed Data Sampling Regression for Long Multi-Horizon Time Series Forecasting}

\newcommand{\TheModel}{DMIDAS }
\newcommand{\TheModelsp}{DMIDAS}
\newcommand\btheta{\boldsymbol \theta}

\begin{document}

\twocolumn[
\icmltitle{\TheModelsp: Deep Mixed Data Sampling Regression for Long Multi-Horizon Time Series Forecasting}




\icmlsetsymbol{equal}{*}

\begin{icmlauthorlist}
\icmlauthor{Cristian Challu}{cmu}
\icmlauthor{Kin G. Olivares}{cmu}
\icmlauthor{Gus Welter}{cmu}
\icmlauthor{Artur Dubrawski}{cmu}
\end{icmlauthorlist}

\icmlaffiliation{cmu}{Auton Lab, School of Computer Science, Carnegie Mellon University}

\icmlcorrespondingauthor{Cristian Challu}{cchallu@cs.cmu.edu}
\icmlcorrespondingauthor{Kin G. Olivares}{kdgutier@cs.cmu.edu}

\icmlkeywords{Deep Learning, Interpretable Neural Network, Time Series Decomposition, Long Horizon Forecasting, Interpolation}

\vskip 0.3in
]



\printAffiliationsAndNotice{} 

\begin{abstract}
Neural forecasting has shown significant improvements in the accuracy of large-scale systems, yet predicting extremely long horizons remains a challenging task. Two common problems are the volatility of the predictions and their computational complexity; we addressed them by incorporating smoothness regularization and mixed data sampling techniques to a well-performing multi-layer perceptron based architecture (NBEATS). We validate our proposed, \TheModelsp, on high-frequency healthcare and electricity price data with long forecasting horizon ($\sim 1000$ timestamps) where we improve the prediction accuracy by 5\% over state-of-the-art models, reducing the number of parameters of NBEATS by nearly 70\%.
\end{abstract}

\section{Introduction} \label{section1:introduction}
Recently neural forecasting has shown great success on improving the accuracy of forecasting systems. Long-horizon forecasting remains a challenging for neural networks as often times their expressiveness translates into excessive computational complexity and volatility. We address these limitations, improving on well performing multi-horizon models with temporal mixed data sampling techniques and smoothness regularization. Our contributions include:
\begin{enumerate}[label=(\roman*)]
    \item \textbf{Mixed Data Sampling}: We incorporate sub-sampling layers before fully-connected networks, and observe that this technique significantly reduces the memory footprint and the amount of computation, while maintaining the effective memory of the model.
    \item \textbf{Smoothness Regularization}: We induce smoothness of the multi-horizon model's predictions by reducing the dimension of its outputs and matching the forecast with the original frequency through interpolation. We add L1 regularization to shrink the weights towards a sparse representation, which robustifies the model against the data's higher frequency noise.
    \item \textbf{\TheModel architecture}: The two regularization techniques naturally motivate the \emph{Deep Mixed Data Sampling} regression (\TheModelsp), that improves the forecast decomposition capabilities of NBEATS by specializing the blocks of the architecture on different frequencies of the data, reducing its volatility, and computational complexity, while maintaining its predictive power.
    
\end{enumerate}

\begin{figure}[tb]
\centering
\subfigure[Arterial Pressure]{\label{fig:a}
\includegraphics[width=0.9\linewidth]{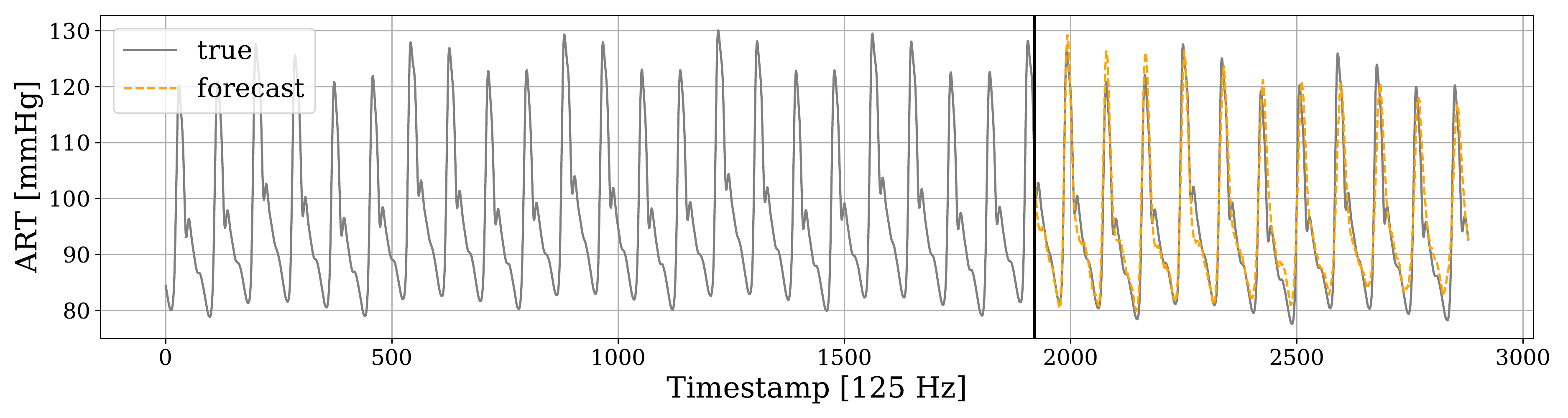}}
\subfigure[French European Power Exchange Electricity Price]{\label{fig:b}
\includegraphics[width=0.9\linewidth]{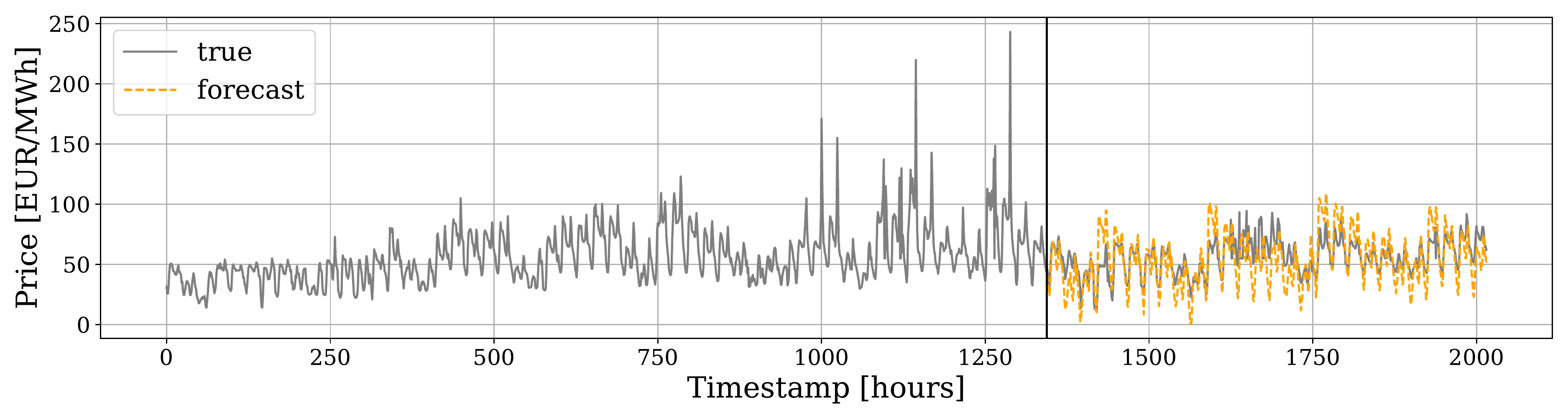}}
\caption{Arterial pressure (ART) and French European Power Exchange electricity price, along with the long-multi-horizon forecasts of \TheModelsp. High-Frequency data often poses challenging forecasting tasks when series display heterogeneous behavior across series and signals have non-stationary dynamics.}
\label{fig:data}
\end{figure}

We compare \TheModel on long horizon forecasting tasks against well-established benchmarks: a parsimonious fully-connected network (MLP; \citealt{lago2021epftoolbox}), the \emph{Dilated Recurrent Neural Network} (DilRNN; \citealt{chang2017dilatedRNN}), the \emph{Exponential Smoothing Recurrent Neural Network} (ESRNN; \citealt{smyl2019esrnn}) and the \emph{Neural Basis Expansion Analysis} (NBEATS; \citealt{oreshkin2020nbeats}). \TheModel reduced RMSE by 3\% and MAE by 5\% against 2nd best model. We improve the average RMSE across datasets on 6\%, 11\%, 10\% over NBEATS-G, NBEATS-I and ESRNN respectively.


The remainder of the work is structured as follows. Section \ref{section2:literature} reviews relevant literature, Section \ref{section3:model} introduces notation and describes the methodology, Section \ref{section4:experiments} contains the empirical findings. Finally in Section \ref{section5:conclusion} we wrap up and conclude.

\section{Literature Review} \label{section2:literature}
Neural forecasting methods have become an increasingly active area of research in recent years, as these models have been successfully adopted in multiple domains such as demand forecasting \citep{wen2017mqrcnn, salinas2020deepAR}, weather prediction \citep{nascimento2019weather} energy markets \citep{olivares2021nbeatsx, gasparin2019electric_load} and excelled at several forecasting competitions \citep{smyl2019esrnn, oreshkin2020nbeats, lim2020temporal_fusion_transformers}. For a comprehensive survey of neural forecasting see \citep{benidis2020dl_timeseries_review2}.

Regarding our approach to tackle the long horizon forecasting task we found the research over multi-step-ahead forecasting strategies and mixed data sampling regressions to be the most relevant, we summarize below.

\textbf{Multi-step-ahead forecasting strategies.} Comprehensive investigations on the bias and variance behavior of multi-step-ahead forecasting strategies observed that the \emph{direct} strategy has a low bias and high variance, with the benefit of avoiding error accumulation across the horizon exhibit by classic \emph{recursive} strategies. Finally, variants of the \emph{joint} or \emph{multi-horizon} strategy allow for the best trade off of the variance and bias effects because they maintain great expressiveness while still sharing parameters \citep{bao2014msvr, atiya2016multi_step_forecasting, wen2017mqrcnn}. 


\textbf{Mixed data sampling regression.} Previous forecasting literature recognized challenges of extremely long horizon predictions, and proposed \emph{mixed data sampling regression} (MIDAS) to ameliorate the problem of parameter proliferation while preserving high frequency temporal information \citep{ghysels2007midas_regressions, armesto2010ForecastingWMF}. MIDAS regressions maintained the classic \emph{recursive} forecasting strategy of linear auto-regressive models, but defined a parsimonious fashion of feeding the inputs to the model.

\textbf{Smoothness Regularization.} Interpolation techniques to augment the resolution of modeled signals has a very long tradition \citep{meijering2002interpolation_history}, with applications in many fields like signal and image processing. In time series forecasting it has many applications, from completing unevenly sampled data and noise filters, to temporal hierarchical forecasting \cite{chow1971interpolation_extrapolation_time_series, roque1981interpolation_time_series_note}. For deep learning, interpolation has seen use in computer vision applications \citep{noh2015deconvolution_network, ye2018convolutional_framelets_inverse}.

To our knowledge, temporal interpolation has not been explicitly leveraged in neural forecasting model's architectures to induce smoothness of its predictions.

\section{Methodology} \label{section3:model}
\subsection{Neural Basis Expansion Analysis}
The \emph{Neural Basis Expansion Analysis} (NBEATS) is a state-of-the-art deep learning univariate model. The key idea of the model is to perform local nonlinear projections onto basis functions across multiple blocks. Each block consists of a fully-connected neural network (MLP) which learns coefficients for the backcast and forecast outputs on a predefined basis. The backcast output is used to clean the inputs of subsequent blocks, while the forecasts are summed to compose the final prediction. The blocks are grouped in stacks, each specialized in learning a different characteristic of the data using different basis functions.

\begin{figure*}[tb]
\centering
\includegraphics[width=0.75\linewidth]{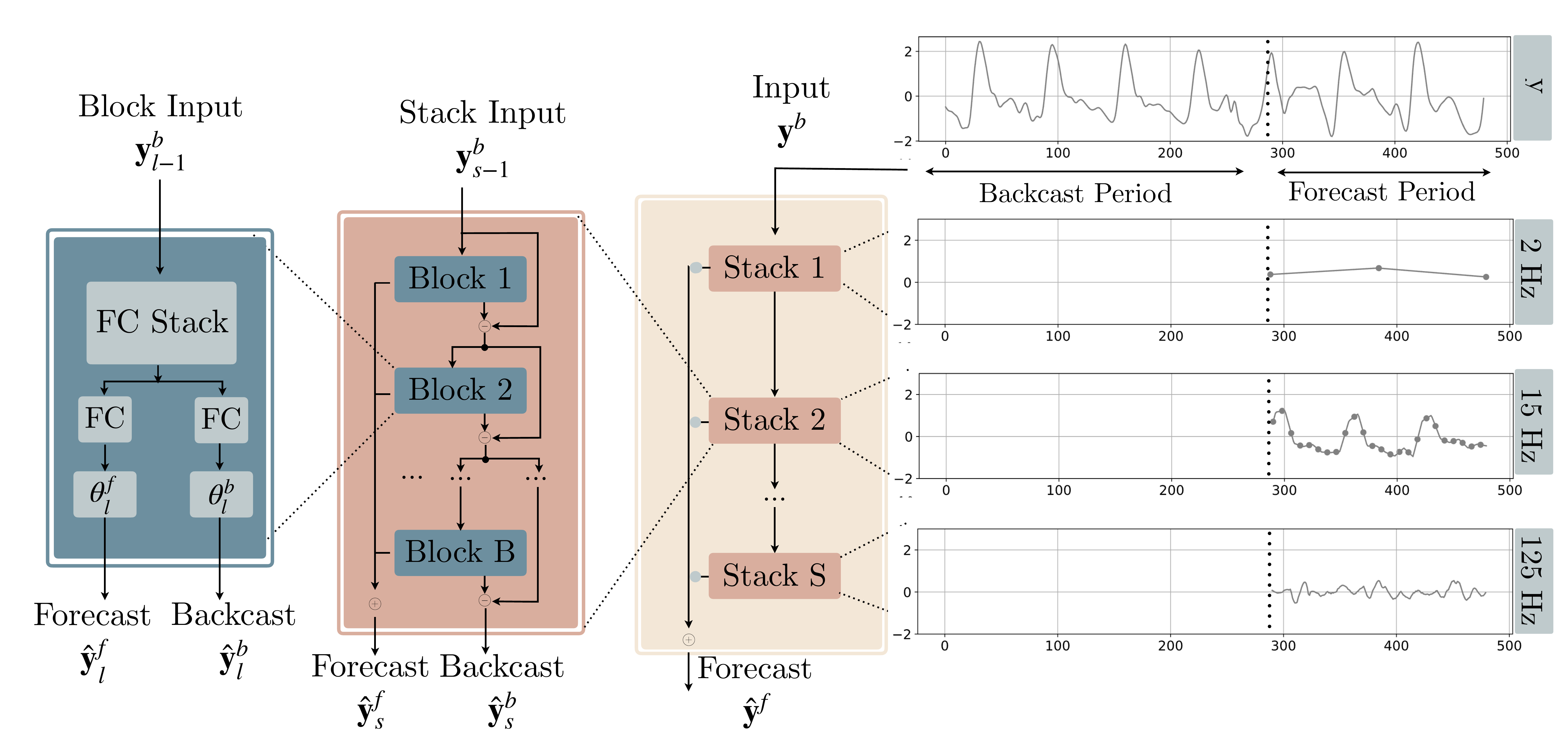}
\caption{Building blocks of the \TheModel architecture. The model is composed several multilayer fully connected networks with ReLU nonlinearities. Blocks overlap using the doubly residual stacking principle for the backcast $\mathbf{\hat{y}}^{b}_{l}$ and forecast $\mathbf{\hat{y}}^{f}_{l}$ outputs of the $l$-th block. The expressiveness of the output dimension of each stack guides the specialization of the additive predictions on different frequencies, while the final predictions are constructed using interpolation.} \label{fig:dmidas_architecture}
\end{figure*}

The NBEATS architecture is composed of $S$ stacks with $B$ blocks each. The input $\mathbf{y}^{b}$ of the first block consists of $L$ lags of the target time-series $\mathbf{y}$, while the inputs of the following blocks include residual connections with the backcast output of the previous block. Within each $l$-th block, the first component consists of a MLP that learns hidden vector $\mathbf{h}_{l}$, which is then passed to a linear layer to produce $\mathbf{\theta}^{f}_{l}$ and backcast $\mathbf{\theta}^{b}_{l}$ expansion coefficients. After the MLP, each $l$-th block includes a \textit{basis expansion operation} between the coefficients learnt and the block's basis function. This transformation results in the backcast $\mathbf{\hat{y}}^{b}_{l}$ and forecast $\mathbf{\hat{y}}^{f}_{l}$ outputs of the block. NBEATS' block operations are:
\begin{subequations} \label{equation:nbeats_projections}
\begin{alignat}{3}
   \mathbf{h}_{l}  &= \mathbf{MLP}_{l}\left(\mathbf{y}^{b}_{l-1}\right) 
   & & & \\
   \btheta^{f}_{l} &= \textbf{LINEAR}^{f}\left(\mathbf{h}_{l}\right)
   & \hspace{1.5em} & \hspace{1.5em} & 
   \btheta^{b}_{l} &= \textbf{LINEAR}^{b}\left(\mathbf{h}_{l}\right) \\
   \mathbf{\hat{y}}^{f}_{l} &= \btheta^{f}_{l} \mathbf{V}^{f}_{l}
   & & & 
   \mathbf{\hat{y}}^{b}_{l} &= \btheta^{b}_{l} \mathbf{V}^{b}_{l}
\end{alignat}
\end{subequations}
For the original \emph{interpretable} model NBEATS-I, a polynomial basis is used to model trends, and harmonic functions to model seasonalities. A block with no inductive bias that directly uses the coefficients $\boldsymbol{\theta}$ as forecast and backcast, i. e. $\mathbf{V}^{f}_{l}=I_{H \times H}$, is used in the \emph{generic} NBEATS-G version.

\subsection{Deep Mixed Data Sampling Regression}

\subsubsection{Mixed Data Sampling}\label{sec:pooling}

In order to capture long time dynamics while not over-parametrizing the model we added pooling layers\footnote{The pooling layers can be average pooling, max pooling or simple stride down sampling.}. Different kernel sizes induce mixed frequencies of the inputs on the backcast window. These sub-sampling layers effectively reduce the number of parameters, limiting the memory footprint and the amount of computation, while maintaining the original receptive field. This technique was previously explored in the MQ-CNN architecture \citep{wen2017mqrcnn}.

\subsubsection{Smoothness Regularization}\label{sec:reg}

For most multi-horizon forecasting models, and in particular NBEATS as specified in Equation~(\ref{equation:nbeats_projections}), the outputs' cardinality corresponds to the horizon's dimension, $|\btheta^{f}_{l}| = H$. As the dimension of the forecast horizon grows, so the model prediction's volatility. To solve this issue, \TheModel defines the dimensions of its forecast coefficients in terms of the \emph{expressivity ratio} $r_{l}$ that controls the number of parameters per unit of time, now $|\btheta^{f}_{l}|= \lceil r_{l} \, H \rceil$. To recover the target horizon $H$ on the original frequency \TheModel uses temporal interpolation\footnote{Other interpolation techniques can be used for example nearest neighbors, or polynomials. In this work we use linear interpolation.}:
\begin{equation}
    \hat{y}^{f}_{l,t} 
    = \left(\theta^{f}_{l,t_{1}} + \left(\frac{\theta^{f}_{l,t_{2}}-\theta^{f}_{l,t_{1}}}{\frac{H}{\lceil r_{l}H \rceil}}\right)(t-t_{1})\right)
    \label{equation:interpolation_projections}
\end{equation}

In Equation~(\ref{equation:interpolation_projections}), $t_{2}$ denotes the closest available time index in the future of $t$ where the coefficients $\btheta^{f}_{l}$ have an associated value, analogous to $t_{1}$ for the past. A consequence of the temporal interpolation, is that the predictions are continuous and smooth between each coefficient $\btheta^{f}_{t_{1}}$ and $\btheta^{f}_{t_{2}}$.

\begin{figure*}[ht!]
\centering
\subfigure[NBEATS-G]{\label{fig:decomposition_g}
\includegraphics[width=0.4\linewidth]{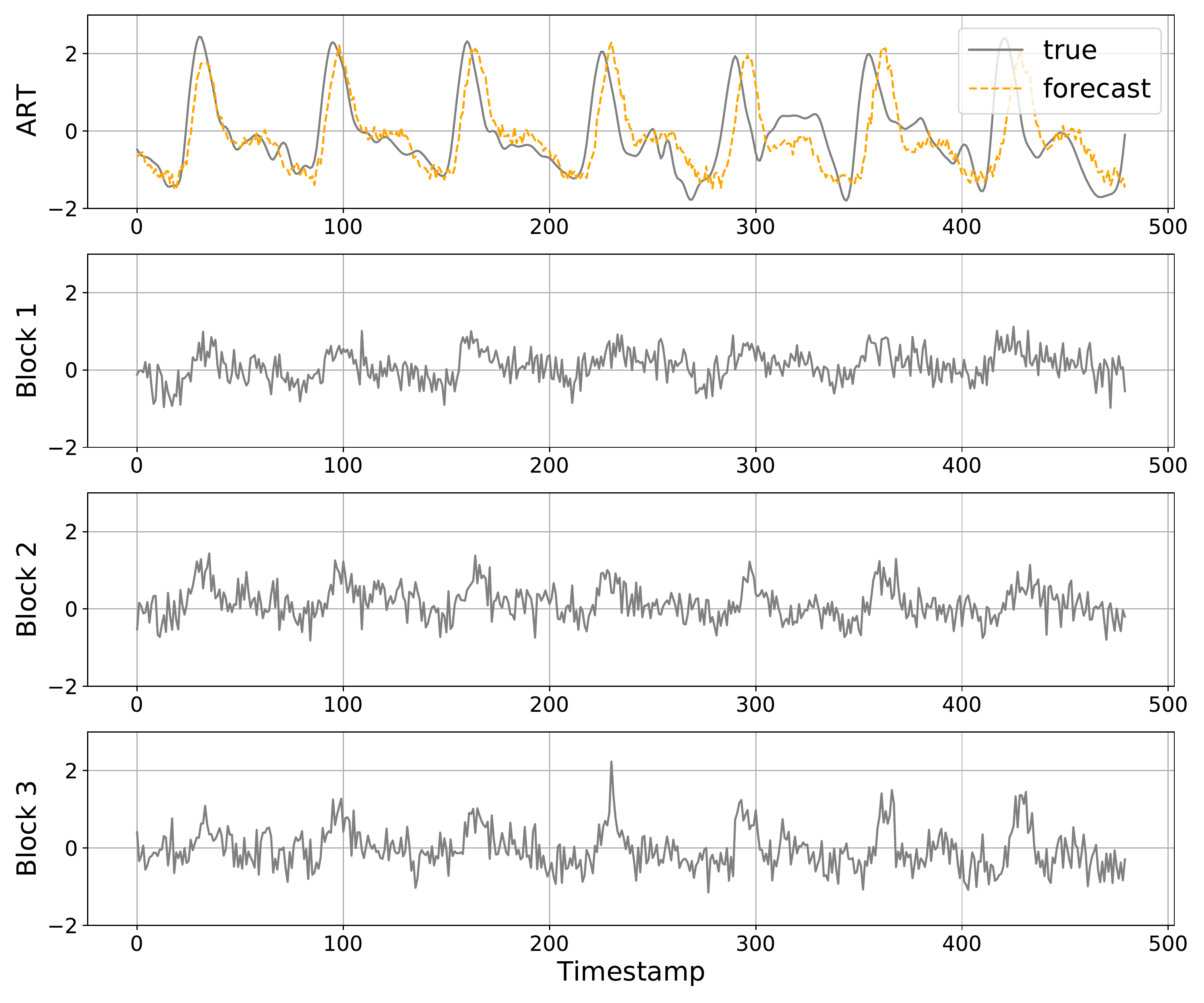}}
\subfigure[\TheModelsp]{\label{fig:decomposition_mf}
\includegraphics[width=0.4\linewidth]{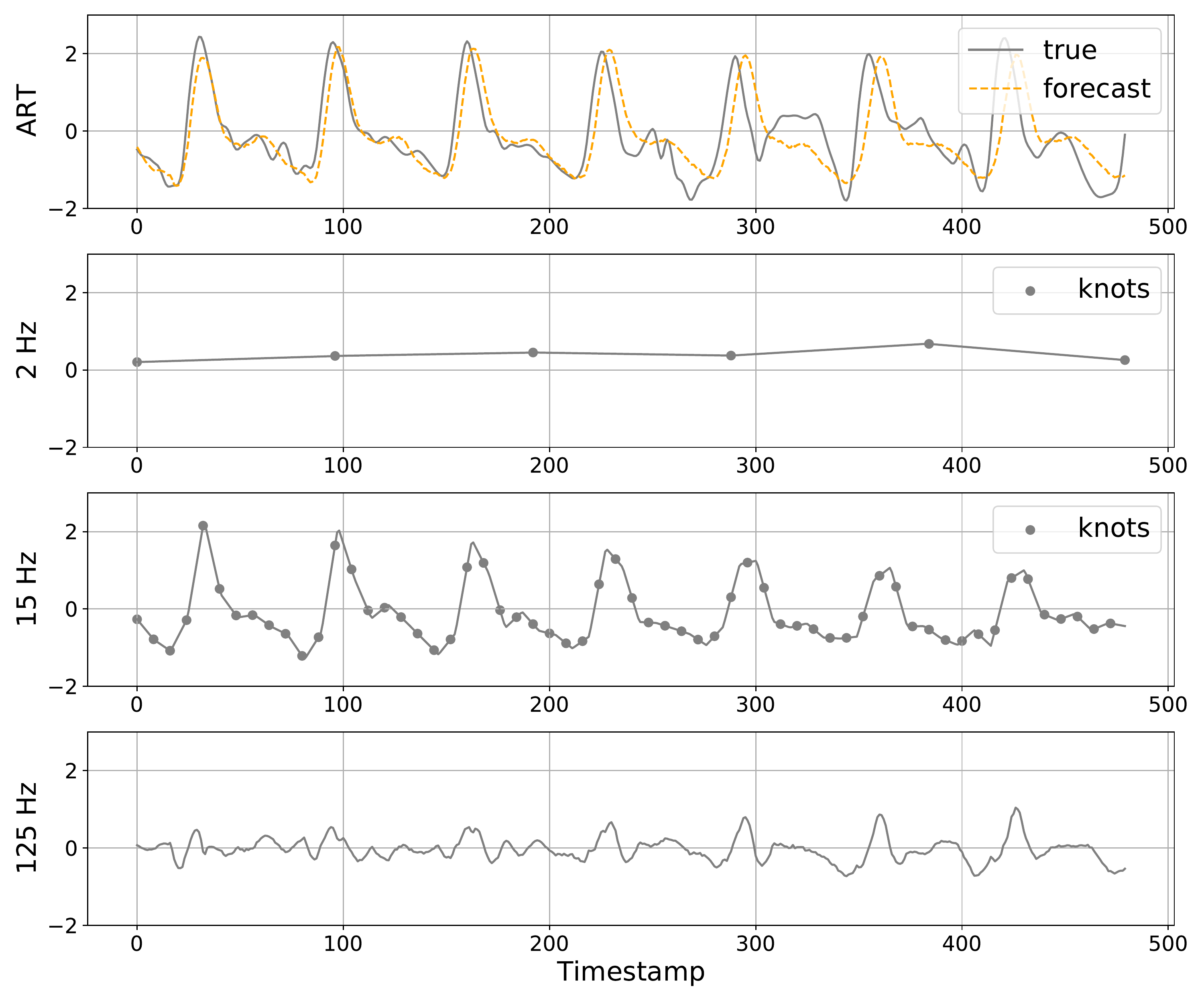}}
\caption{Arterial Pressure (ART) and five hundred steps ahead forecasts using NBEATS-G and \TheModelsp. The top row shows the original signal. The second, third and fourth rows show the forecast components for the each model's block, in the case of sub-Figure (b) each block specializes on different frequencies, contrary to the outputs of NBEATS-G on sub-Figure (a) that are unintelligible.} \label{fig:decomposition}
\end{figure*}

\subsubsection{\TheModel Architecture}\label{sec:multi_freq}

The ideas from multi-horizon forecasting, mixed data sampling regressions and smoothness regularization through interpolation naturally converge into \TheModel approach. We design the architecture so that each block specializes on different sampling frequencies of the inputs and outputs of the model simultaneously, by selecting the \emph{expressivity ratio}. We define the coefficients $\btheta^{f}_{l}$ and $\btheta^{b}_{l}$ for each $l$-th block to be uniformly spaced, as shown in Figure~\ref{fig:dmidas_architecture}.

We use \emph{exponentially increasing expressivity} ratios through the depth of the architecture blocks that allows to model complex dependencies, while controlling the number of parameters used on each output layer. If the \emph{expressivity ratio} is defined as $r_{l} = r^{l}$ then the space complexity of \TheModel scales geometrically $\mathcal{O}\left((H(1-r^{B})/(1-r)\right)$, while the NBEATS-G scales linearly $\mathcal{O}\left(H B \right)$.

\TheModel enjoys several advantages. First, it reduces the volatility of the model's outputs and improves the overall computational efficiency. Second, the additive decomposition capabilities inherited from NBEATS improves by making each block specialize on different frequencies. Furthermore, the expressivity ratio $r_l$ can be specified using the generating process domain expertise to improve forecasting performance. Third, L1 regularization can continue to shrink the weights towards a sparse representation, robustifying the model to higher frequency noise.

The additive forecast decomposition of \TheModel provides valuable information beyond the trend-seasonality decomposition. Figure \ref{fig:decomposition} shows a forecast of \TheModel and its corresponding decomposition for ART and compares it with the non-interpretable decomposition of the NBEATS-G.

\section{Experiments} \label{section4:experiments}

\textbf{Vital Signs Dataset.} This dataset consists of de-identified high-frequency vital signs collected from the intensive-care at the University of Pittsburgh Medical Center Hospitals (UPMC) over three years \footnote{The data was collected from Phillips Data Warehouse Connect, and it was prepared and de-identified under Institutional Review Board review and approval}. It contains \emph{arterial blood pressure} (ART) and \emph{Pulse Oximetry Photoplethysmogram} (PLETH) waveforms for 98 patients. Each patient has 90-minutes data, the last five minutes comprise the test set.

\textbf{Electricity Price Dataset.} We consider \emph{electricity price forecasting} (EPF) datasets of five major power markets \footnote{Available at the \href{https://github.com/jeslago/epftoolbox}{EPFtoolbox library} \citep{lago2021epftoolbox}.}, namely Nord Pool, the Pennsylvania-New  Jersey-Maryland market, and the European Power Exchange for Belgium, France and Germany. Each market contains six years of history, we keep the last three months of each as test set.

\subsection{Training Methodology and Evaluation}

For ART and PLETH we train global models. For EPF we train separate models for each market. A simple mean ensemble of four random initializations is used to forecast. Model selection is done with a bayesian optimization technique which explores the hyperparameter space using tree-structured Parzen estimators~(HYPEROPT; \citealt{bergstra2011hyperopt}). The configurations that reach the lowest validation \emph{mean absolute error} (MAE) are evaluated on the test data. 

We evaluate the accuracy with the \emph{mean absolute error} (MAE) and \emph{root mean squared error} (RMSE). \TheModel consistently achieved the best performance on long horizons, with a monotonic increasing relative improvement vs horizon. It improves the RMSE \textbf{3\%} on average and MAE on \textbf{5\%} against the 2nd best model, including several models and forecasting strategies, such as the recursive forecast of the DilRNN, while remaining interpretable. \TheModel improve the average RMSE across datasets on \textbf{6\%}, \textbf{11\%}, \textbf{10\%} over NBEATS-G, NBEATS-I and ESRNN respectively.

\begin{table}[ht]
\label{table:main_result}
\centering
\caption{Forecast accuracy measures for long-horizon tasks. The reported metrics are \emph{mean absolute error} (MAE) and \emph{root mean squared error} (RMSE). Smallest errors are highlighted in bold.}
\resizebox{\columnwidth}{!}{%
\begin{tabular}{lll|cccccc}
            \toprule
            \textbf{Data} & \textbf{H} & \textbf{Metric}  & \textbf{MLP}  & \textbf{DilRNN} & \textbf{ESRNN} & \textbf{NBEATS-I} & \textbf{NBEATS-G} & \textbf{\TheModelsp}\\ \hline
\multirow{8}{*}{ART} & \multirow{2}{*}{120} & RMSE & 13.17 & 13.03 & 12.48 & \textbf{11.96} & 12.28 & 12.40 \\
&             & MAE & 6.33 & 6.62 & 5.89 & \textbf{5.25} & 5.30 & 5.44  \\\cmidrule{2-9}
& \multirow{2}{*}{480} & RMSE & 18.17 & 17.88 & 22.40 & 17.48 & 16.93 & \textbf{16.71} \\
&             & MAE & 9.21 & 9.12 & 13.15 & 8.73 & 7.80 & \textbf{7.68}  \\\cmidrule{2-9}
& \multirow{2}{*}{960} & RMSE & 21.24 & 21.20 & 21.87 & 22.11 & 18.64 & \textbf{18.01} \\
&             & MAE & 12.23 & 11.84 & 12.04 & 14.17 & 9.97 & \textbf{9.87} \\
           \midrule
           \midrule
\multirow{8}{*}{PLETH} & \multirow{2}{*}{120} & RMSE & 0.054 & 0.056 & 0.060 & 0.051 & 0.050 & \textbf{0.050}\\
&             & MAE & 0.033 & 0.034 & 0.035 & 0.029 & \textbf{0.028} & \textbf{0.028} \\\cmidrule{2-9}
& \multirow{2}{*}{480} & RMSE & 0.067 & 0.071 & 0.106 & 0.065 & 0.063 & \textbf{0.061} \\
&             & MAE & 0.045 & 0.046 & 0.074 & 0.043 & \textbf{0.039} & 0.041 \\\cmidrule{2-9}
& \multirow{2}{*}{960} & RMSE & 0.079 & 0.081 & 0.078 & 0.082 & \textbf{0.073} & \textbf{0.073} \\
&             & MAE & 0.054 & 0.055 & 0.058 & 0.059 & 0.049 & \textbf{0.046} \\
           \midrule
           \midrule
\multirow{8}{*}{EPF} & \multirow{2}{*}{24} & RMSE & 9.84 & 9.66 & 9.55 & 9.49 & 9.34 & \textbf{9.04} \\
&  & MAE & 6.02 & 5.90 & 5.81 & 5.89 & 5.65 & \textbf{5.56} \\\cmidrule{2-9}
& \multirow{2}{*}{336} & RMSE  & 12.94 & 12.99 & \textbf{12.84} & 13.29 & 13.00 & \textbf{12.84} \\
&             & MAE & 8.80 & 8.80 & 8.75 & 9.05 & 8.76 &  \textbf{8.74} \\\cmidrule{2-9}
& \multirow{2}{*}{672} & RMSE & 18.03 & 17.45 & 17.33 & 16.92 & 18.11 & \textbf{15.88} \\\
&  & MAE & 13.30 & 13.32 & 13.01 & 12.61 & 13.43 & \textbf{11.50} \\
\bottomrule
\end{tabular}
}
\end{table}

\section{Conclusion} \label{section5:conclusion}
We identified two challenges in long-horizon forecasting tasks, namely the volatility of multi-horizon model's predictions and their computational complexity. We proposed the \emph{Deep Mixed Data Sampling regression} (\TheModelsp), that incorporates smoothness regularization through interpolation and sub-sampling techniques to NBEATS. The resultant parsimonious model outperforms state-of-the-art benchmarks on long-horizon forecasting tasks, improving the MAE 5\% on average and the RMSE 3\%, while reducing the numbers of parameters of the NBEATS model by nearly 70\%. Additionally, \TheModel has an interpretable forecast decomposition that provides valuable information beyond the classic trends and seasons.

\clearpage
\bibliography{citations}
\bibliographystyle{icml2021}

\appendix

\end{document}